%% file: main.tex
\documentclass{article}

\usepackage{microtype}
\usepackage{graphicx}
\usepackage{subfigure}
\usepackage{booktabs} % for professional tables
\usepackage{amsmath}
\usepackage{amssymb}
\usepackage{bbm}
\usepackage{amsfonts}
\usepackage{todonotes}
\usepackage{algorithm}
\usepackage{enumitem}
\usepackage{multirow}
\usepackage{stfloats}
\usepackage{parskip}
\include{my_defs}

\usepackage{hyperref}
\usepackage{cleveref}

\usepackage{stfloats}

\usepackage{fullpage}
\usepackage[round]{natbib}

\title{Representation Matters: Offline Pretraining\\for Sequential Decision Making}
\author{
 \begin{tabular}{p{6cm}p{6cm}}
  {\hfil Mengjiao Yang}& {\hfil Ofir Nachum} \\[.1cm]
  {\hfil \texttt{sherryy@google.com}}
    & {\hfil\texttt{ofirnachum@google.com}} \\[.2cm]    
    \hfil \normalfont Google Research & \hfil \normalfont Google Research \\[0.1cm]
  \end{tabular} 
}
\date{}
  
\begin{document}
\maketitle

\begin{abstract}
\noindent
The recent success of supervised learning methods on ever larger offline datasets has spurred interest in the reinforcement learning (RL) field to investigate whether the same paradigms can be translated to RL algorithms. 
This research area, known as \emph{offline RL}, has largely focused on offline policy optimization, aiming to find a return-maximizing policy exclusively from offline data. 
In this paper, we consider a slightly different approach to incorporating offline data into sequential decision-making. 
We aim to answer the question, what unsupervised objectives applied to offline datasets are able to learn state representations which elevate performance on downstream tasks, whether those downstream tasks be online RL, imitation learning from expert demonstrations, or even offline policy optimization based on the same offline dataset? 
Through a variety of experiments utilizing standard offline RL datasets, we find that the use of pretraining with unsupervised learning objectives can dramatically improve the performance of policy learning algorithms that otherwise yield mediocre performance on their own.
Extensive ablations further provide insights into what components of these unsupervised objectives -- e.g., reward prediction, continuous or discrete representations, pretraining or finetuning -- are most important and in which settings.
\end{abstract}

\input{intro}
\input{related}

\input{background}
\input{experiment}
\input{conc}

\section*{Acknowledgements}
We thank Ilya Kostrikov for providing his codebase as a starting point for our experiments as well as helpful comments on early drafts of this work. We also thank the rest of the Google Brain team for insightful thoughts and discussions.

\bibliography{main}
\bibliographystyle{plainnat}
\input{appendix}

\end{document}

%% file: my_defs.tex
\usepackage{mathtools}

\newcommand{\E}{\mathop{\mathbb{E}}}

\newcommand{\secref}[1]{Section~\ref{#1}}

\newcommand{\appref}[1]{Appendix~\ref{#1}}

\newcommand{\tabref}[1]{Table~\ref{#1}}
\newcommand{\figref}[1]{Figure~\ref{#1}}

\def\E{\mathbb{E}}

%% file: intro.tex
% !TEX root = main.tex
\section{Introduction}
Within the reinforcement learning (RL) research field, \emph{offline RL} has recently gained a significant amount of interest~\citep{levine2020offline,lange2012batch}.
Offline RL considers the problem of performing reinforcement learning -- i.e., learning a policy to solve a sequential decision-making task -- exclusively from a static, offline dataset of experience.
The recent interest in offline RL is partly motivated by the success of \emph{data-driven} methods in the supervised learning literature. 
Indeed, the last decade has witnessed ever more impressive models learned from ever larger static datasets~\citep{halevy2009unreasonable,krizhevsky2012imagenet,brown2020language,dosovitskiy2020image}.
Solving offline RL is therefore seen as a stepping stone towards developing scalable, data-driven methods for policy learning~\citep{fu2020d4rl}.
Accordingly, much of the recent offline RL research focuses on proposing new policy optimization algorithms amenable to learning from offline datasets (e.g.,~\citet{fujimoto2019off,wu2019behavior,agarwal2020optimistic,kumar2020conservative,yu2020mopo,matsushima2020deployment}).

In this paper, we consider a slightly different approach to incorporating offline data into sequential decision-making. 
We are inspired by recent successes in semi-supervised learning~\citep{mikolov2013efficient,devlin2018bert,chen2020simple}, in which large and potentially unlabelled offline datasets are used to learn \emph{representations} of the data -- i.e., a mapping of input to a fixed-length vector embedding -- and these representations are then used to accelerate learning on a downstream supervised learning task.
We therefore consider whether the same paradigm can apply to RL. Can offline experience datasets be used to learn representations of the data that accelerate learning on a downstream task?

\begin{figure}
    \centering
    \includegraphics[width=.8\linewidth]{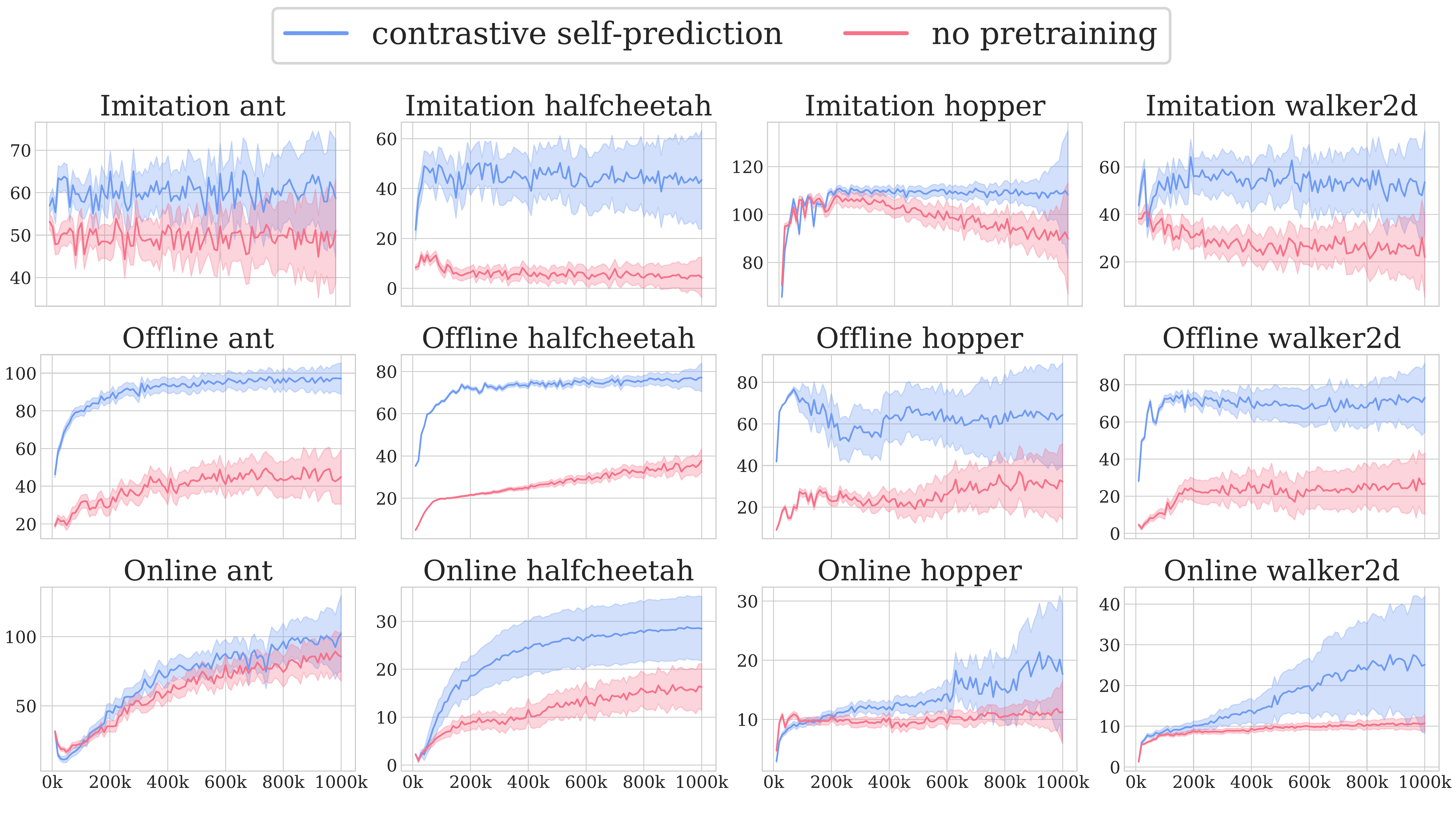}
    \caption{A summary of the advantages of representation learning via contrastive self-prediction, across a variety of settings: imitation learning, offline RL, and online RL. 
    Each subplot shows the aggregated mean reward and standard error during training, with aggregation over offline datasets of different behavior (e.g., expert, medium, etc.), with five seeds per dataset (see~\secref{sec:setup}). Representation learning yields significant performance gains in all domains and tasks.}
    \label{fig:line_plot}
\end{figure}

This broad and general question has been partially answered by previous works~\citep{ajay2020opal,singh2020parrot}.
These works focus on using offline datasets to learn representations of \emph{behaviors}, or actions. More specifically, these works learn a spectrum of behavior policies, conditioned on a latent $z$, through supervised action-prediction on the offline dataset. 
The latent $z$ then effectively provides an abstract action space for learning a hierarchical policy on a downstream task, and this straightforward paradigm is able to accelerate learning in a variety of sequential decision-making settings.
Inspired by these promising results and to differentiate our own work, we focus our efforts on the question of representation learning for \emph{observations}, or states, as opposed to learning representations of behaviors or actions.
That is, we aim to answer the question, can offline experience datasets be used to learn representations \emph{of state observations} such that learning policies from these pretrained representations, as opposed to the raw state observations, improves performance on a downstream task?\footnote{Whether the two aspects of representation learning -- action representations and state representations -- can be combined is an intriguing question. However, to avoid an overly broad paper, we focus only on state representation learning, and leave the question of combining this with action representation learning to future work.}

To approach this question, we devise a variety of offline datasets and corresponding downstream tasks. For offline datasets, we leverage the Gym-MuJoCo datasets from D4RL~\citep{fu2020d4rl}, which provide a diverse set of datasets from continuous control simulated robotic environments.
For downstream tasks, we consider three main categories: (1) low-data imitation learning, in which we aim to learn a task-solving policy from a small number of expert trajectories; (2) offline RL, in which we aim to learn a task-solving policy from the same offline dataset used for representation learning; and (3) online RL, in which we aim to learn a task-solving policy using online access to the environment.

Once these settings are established, we then continue to evaluate the ability of state representation learning on the offline dataset to accelerate learning on the downstream task.
Our experiments are separated into two parts, \emph{breadth} and \emph{depth}. 
First for breadth, we consider a diverse variety of representation learning objectives taken from the RL and supervised learning literature. The results of these experiments show that, while several of these objectives perform poorly, a few yield promising results.
This promising set essentially comprises of objectives which we call \emph{contrastive self-prediction}; these objectives take sub-trajectories of experience and then use some components of the sub-trajectory to predict other components, with a contrastive loss when predicting states -- e.g., using a contrastive loss on the affinity between a sequence of states and actions and the next state, akin to popular methods in the supervised learning literature~\citep{mikolov2013efficient, devlin2018bert}.

These initial findings guide our second set of experiments. 
Aiming for depth, we devise an extensive ablation based on contrastive self-prediction to investigate what components of the objective are most important and in which settings. 
For example, whether it is important to include reward as part of the sub-trajectory, or whether discrete representations are better than continuous, whether pre-training and fixing the representations is better than finetuning, etc.
In short, we find that state representation learning can yield a dramatic improvement in downstream learning. 
Compared to performing policy learning from raw observations, we show that relatively simple representation learning objectives on offline datasets can enable better and faster learning on imitation learning, offline RL, and online RL (see Figure~\ref{fig:line_plot}).
We believe these results are especially compelling for the imitation learning setting -- where even a pretraining dataset that is far from expert behavior yields dramatic improvement in downstream learning -- and in the offline RL setting -- where we show the benefits of representation learning are significant even when the pretraining dataset \emph{is the same as} the downstream task dataset.
We hope that these impressive results guide and encourage future researchers to develop even better ways to incorporate representation learning into sequential decision-making.

%% file: related.tex
% !TEX root = main.tex
\section{Background and Related Work}
\label{sec:related}

Representation learning for RL has a rich and diverse existing literature, and we briefly review these relevant works. 

\paragraph{Abstraction and Bisimulation}
Traditionally, representation learning has been framed as learning or identifying \emph{abstractions} of the state or action space of an environment~\citep{andre2002state,mannor2004dynamic,dearden1997abstraction,abel2018state}.
These methods aim to reduce the original environment state and action spaces to more compact spaces by clustering those states and actions which yield similar rewards and dynamics.
Motivated by similar intuitions, research into \emph{bisimulation} metrics has aimed to devise or learn similarity functions between states~\citep{ferns2004metrics,castro2010using}.
While these methods originally required explicit knowledge of the reward and dynamics functions of the environment, a number of recent works have translated these ideas to stochastic representation learning objectives using deep neural networks~\citep{gelada2019deepmdp,zhang2020learning,agarwal2021contrastive}.
Many of these modern approaches effectively learn reward and transition functions in the learned embedding space, and training of these models is used to inform the learned state representations.

\paragraph{Representations in Model-Based Learning}
The idea of learning latent state representations via learning reward and dynamics models leads us to related work in the model-based RL literature.
Several recent model-based RL methods use latent state representations as a way to simplify the model learning and policy rollout elements of model-based policy optimization~\citep{oh2017value,silver2018general,hafner2020mastering}, with the rollout in latent space sometimes referred to as `imagination'~\citep{racaniere2017imagination,hafner2019dream}.
Similar ideas have also appeared under the label of `embed to control'~\citep{watter2015embed,levine2019prediction}.
Other than learning representations through forward models, there are also works which propose to learn \emph{inverse} models, in which an action is predicted based on the representations of its preceding state and subsequent state~\citep{pathak2017curiosity,ShelhamerMAD16}.

\paragraph{Contrastive Objectives}
Beyond model-based representations, many previous works propose the use of contrastive losses as a way of learning useful state representations~\citep{wu2018laplacian,nachum2018near,srinivas2020curl,stooke2020decoupling}.
These works effectively define some notion of similarity between states and use a contrastive loss to encourage similar states to have similar representations.
The similarity is usually based on either temporal vicinity (pairs of states which appear in the same sub-trajectory) or user-specified augmentations, such as random shifts of image observations~\citep{srinivas2020curl}.
Previous work has established connections between the use of contrastive loss and mutual information maximization~\citep{oord2019representation} and energy-based models~\citep{lecun2005loss}.

\paragraph{State Representation Learning in Offline RL}
The existing works mentioned above almost exclusively focus on online settings, often learning the representations on a continuously evolving dataset and in tandem with online policy learning.
In contrast, our work focuses on representation learning on offline datasets and separated from downstream task learning. 
This serves two purposes: First, using static offline datasets makes comparisons between different methods easier, avoiding confounding factors associated with issues of exploration or nonstationary datasets. Second, the offline setting is arguably more practical; in practice, static offline datasets are more common than cheap online access to an environment~\citep{levine2020offline}.
Previous work in a similar vein to ours includes~\citet{stooke2020decoupling} and~\citet{ShelhamerMAD16}, which propose to use unsupervised pretraining, typically only on expert demonstrations, as a way of initializing an image encoder for downstream online RL. 
Our own work complements these existing studies, by presenting extensive comparisons of a variety of representation learning objectives in several distinct settings.
Moreover, our work is unique for showing benefits of representation learning on non-image tasks, thus avoiding the use of any explicit or implicit prior knowledge that is typically exploited for images (e.g., using image-based augmentations or using a convolutional network architecture).

%% file: background.tex
% !TEX root = main.tex
\section{Task Setups}
\label{sec:setup}
We now continue to our own contributions, starting by elaborating on the experimental protocol we design to evaluate representation learning in the context of low-data imitation learning, offline RL (specifically, offline policy optimization), and online RL in partially observable environments.
This protocol is summarized in Table~\ref{tab:proto}.

\begin{table*}[ht]
\begin{center}
\caption{A summary of our experimental setups. In total, there are 16 choices of offline data and downstream task combinations each for imitation learning, offline RL, and online RL. Given that we run each setting with five random seeds, this leads to a total of 240 training runs for every representation learning objective we consider.}
\label{tab:proto}
\setlength{\tabcolsep}{1pt}
\resizebox{\textwidth}{!}{
\begin{tabular}{|lcl|}
    \hline
    & \bf Imitation & \\
    \small Choose $\text{domain}\in\{\text{halfcheetah},\text{hopper},\text{walker2d},\text{ant}\}$ 
    & \multirow{3}{*}{$\rightarrow$} 
    & \multirow{3}{*}{
    \begin{tabular}{rl}
    \small Offline dataset: & \small \{domain\}-\{data\}-v0 \\
    \small Downstream task: & \small Behavioral cloning (BC) on first $N$  \\
    & \small transitions from \{domain\}-expert-v0
    \end{tabular}
    } \\
    \small Choose $\text{data}\in\{\text{medium},\text{medium-replay}\}$ & & \\
    \small Choose $N\in\{10000, 25000\}$ & & \\
    \hline
    & \bf Offline RL & \\
    \small Choose $\text{domain}\in\{\text{halfcheetah},\text{hopper},\text{walker2d},\text{ant}\}$ 
    & \multirow{3}{*}{$\rightarrow$} 
    & \multirow{3}{*}{
    \begin{tabular}{rl}
    \small Offline dataset: & \small \{domain\}-\{data\}-v0 \\
    \small Downstream task: & \small Behavior regularized actor critic (BRAC) \\
    & \small on data from \{domain\}-\{data\}-v0
    \end{tabular}
    } \\
    \small Choose $\text{data}\in\{\text{expert},\text{medium-expert},\text{medium},$ & & \\
    \small \hfill $\text{medium-replay}\}$ & & \\
    \hline
    & \bf Online RL & \\
    \small Choose $\text{domain}\in\{\text{halfcheetah},\text{hopper},\text{walker2d},\text{ant}\}$ 
    & \multirow{3}{*}{$\rightarrow$} 
    & \multirow{3}{*}{
    \begin{tabular}{rl}
    \small Offline dataset: & \small \{domain\}-\{data\}-v0 with random masking \\
    \small Downstream task: & \small Soft actor critic (SAC) on randomly \\
    & \small  masked version of \{domain\}
    \end{tabular}
    } \\
    \small Choose $\text{data}\in\{\text{expert},\text{medium-expert},\text{medium}\}$ & & \\
    \small \hfill $\text{medium-replay}\}$ & & \\
    \hline
\end{tabular}
}
\end{center}
\end{table*}

\subsection{Datasets}
We leverage the Gym-MuJoCo datasets from D4RL~\citep{fu2020d4rl}.
These datasets are generated from running policies on the well-known MuJoCo benchmarks of simulated locomotive agents: halfcheetah, hopper, walker2d, and ant.
Each of these four domains is associated with four datasets -- expert, medium-expert, medium, and medium-replay -- corresponding to the quality of the policies used to collect that data. 
Each dataset is composed of a number of trajectories $\tau:=(s_0,a_0,r_0,s_1,a_1,r_1,\dots,s_T)$.
For example, the dataset ant-expert-v0 is a dataset of trajectories generated by expert task-solving policies on the ant domain, while the dataset halfcheetah-medium-v0 is generated by mediocre, far from task-solving, policies.

Notably, although the underyling MuJoCo environments are Markovian, the datasets are not necessarily Markovian, as they may be generated by multiple distinct policies.

\subsection{Imitation Learning in Low-Data Regime}
Imitation learning~\citep{hussein2017imitation} seeks to match the behavior of an agent with that of an expert. While expert demonstrations are often limited and expensive to obtain in practice, non-expert experience data (e.g., generated from a mediocre agent randomly interacting with an environment) can be much more easily accessible. 

To mimic this practical scenario, we consider an experimental protocol in which the downstream task is behavioral cloning~\citep{pomerleau1991efficient} on a small set of expert trajectories -- selected by taking either the first $10$k or $25$k transitions from an expert dataset in D4RL, corresponding to about 10 and 25 expert trajectories, respectively. 
We then consider either the medium or medium-replay datasets from the same domain for representation learning.\footnote{To avoid issues of extrapolation when transferring learned representations to the expert dataset, we include the small number of expert demonstrations in the offline dataset during pretraining.}
Thus, this set of experiments aims to determine whether representations learned from large datasets of mediocre behavior can help elevate the performance of behavioral cloning on a much smaller expert dataset.

\subsection{Offline RL with Behavior Regularization}

One of the main motivations for the introduction of the D4RL datasets was to encourage research into fully offline reinforcement learning; i.e., whether it is possible to learn return-maximizing policies exclusively from a static offline dataset.
Many algorithms for this setting have recently been proposed, commonly employing some sort of \emph{behavior regularization}~\citep{kumar19bear,jaques2019way,wu2019behavior}. In its simplest form, behavior regularization augments a vanilla actor-critic algorithm with a divergence penalty measuring the divergence of the learned policy from the offline data, thus compelling the learned policy to choose the same actions appearing in the dataset.

While the actor and critic are typically trained with the raw observations as input, with this next set of experiments, we aim to determine whether representation learning can help in this regime as well.
In this setting, the pretraining and downstream datasets are the same, determined by a single choice of domain (halfcheetah, hopper, walker2d, or ant) and data (expert, medium-expert, medium, or medium-replay).
For the downstream algorithm, we use behavior regularized actor-critic (BRAC)~\citep{wu2019behavior}, which is a simple behavior regularized method employing a KL divergence penalty. Notably, although the original BRAC paper uses different regularization strengths and policy learning rates for different domains, we fix these to values which we found to generally perform best (regularization strength of $1.0$ and policy learning rate of $0.00003$).

Thus, this set of experiments aims to determine whether learning BRAC from learned state representations is better (in terms of performance and less dependence on hyperparameters) than learning BRAC from the raw states, even when the state representations are learned using the same offline dataset.

\subsection{Online RL in Partially Observable Environments} 
In this set of experiments, we aim to determine whether representations learned from offline datasets can improve or accelerate learning in an online domain.
One of the most popular online RL algorithms is soft actor critic (SAC)~\citep{haarnoja2019soft}.
SAC is a well-performing algorithm on its own, and so to increase the difficulty of the downstream task, we consider a simple modification to make our domains partially observable: zero-masking out a random dimension of the state observation.
This modification also brings our domains closer to practice, where partial observability due to flaky sensor readings is common~\citep{dulacarnold2019challenges}.

Accordingly, the offline dataset is determined by a choice of domain (halfcheetah, hopper, walker2d, or ant) and data (expert, medium-expert, medium, or medium-replay), with the same masking applied to this dataset.
Representations learned on this dataset are then applied downstream, where SAC is trained on the online domain, with the representation module providing an embedding of the masked observations of the environment within a learned embedding space.

\subsection{Evaluation}
Each representation learning variant we evaluate is run with five seeds on each of the experimental setups described above.
Unless otherwise noted, a single seed corresponds to an initial pretraining phase of $200$k steps, in which a representation learning objective is optimized using batches of 256 sub-trajectories randomly sampled from the offline dataset.
After pretraining, the learned representation is fixed and applied to the downstream task, which performs the appropriate training (BC, BRAC, or SAC) for $1$M steps.
In this downstream phase, every $10$k steps, we evaluate the learned policy on the downstream domain environment by running it for $10$ episodes and computing the average total return.
We normalize this total return according to the normalization proposed in~\citet{fu2020d4rl}, such that a score of 0 roughly corresponds to a random agent and a score of 100 to an expert agent.
We average the last 10 evaluations within the $1$M downstream training, and this determines the final score for the run.
To aggregate over multiple seeds and task setups, we simply compute the average and standard error of this final score.

%% file: experiment.tex
% !TEX root = main.tex
\section{Experiments: Breadth Study}
\label{sec:breadth}
We begin our empirical study with an initial assessment into the performance of a broad set of representation learning ideas from the existing literature. 

\subsection{Representation Learning Objectives}
We describe the algorithms we consider below. While it is infeasible for us to extensively evaluate all previously proposed representation learning objectives, our choice of objectives here aims to cover a diverse set of recurring themes and ideas from previous work  (see Section~\ref{sec:related}).

We use the notation 
\begin{equation*}
\tau_{t:t+k}:=(s_t,a_t,r_t,\dots,s_{t+k-1},a_{t+k-1},r_{t+k-1},s_{t+k})
\end{equation*}
to denote a length-$(k+1)$ sub-trajectory of state observations, actions, and rewards; we use $s_{t:t+k},a_{t:t+k},r_{t:t+k}$ to denote a subselection of this trajectory based on states, actions, and rewards, respectively.
We use $\phi$ to denote the representation function; i.e., $\phi(s)$ is the representation associated with state observation $s$, and $\phi(s_{t:t+k}):=(\phi(s_t),\dots,\phi(s_{t+k}))$.
All learned functions, including $\phi$, are parameterized by neural networks. Unless otherwise noted, $\phi$ is parameterized as a two-hidden layer fully-connected network with 256 units per layer and output of dimension 256 (see further details in~\appref{app:exp}).

\paragraph{Inverse model} Given a sub-trajectory $\tau_{t:t+1}$, use $\phi(s_{t:t+1})$ to predict $a_t$. That is, we train an auxiliary $f$ such that $f(\phi(s_{t:t+1}))$ is a distribution over actions, and the learning objective is $-\log P(a_t | f(\phi(s_{t:t+1})))$. This objective may be generalized to sequences longer than $k+1=2$ as $-\log P(a_{t+k-1} | f(\phi(s_{t:t+k}),a_{t:t+k-1}))$.

\paragraph{Forward raw model} Given a sub-trajectory $\tau_{t:t+1}$, use $\phi(s_t),a_t$ to predict $r_t, s_{t+1}$. That is, we train an auxiliary $f,g$ such that $f(\phi(s_t),a_t)$ is a distribution over next states and $g(\phi(s_t),a_t)$ is a scalar reward prediction. The learning objective is $||r_t-g(\phi(s_t),a_t)||^2 - \log P(s_{t+1} | f(\phi(s_t),a_t))$. This objective may be generalized to sequences longer than $k+1=2$ as $||r_t-g(\phi(s_{t:t+k-1}),a_{t:t+k-1})||^2 - \log P(s_{t+1} | f(\phi(s_{t:t+k-1}),a_{t:t+k}))$.

\paragraph{Forward latent model; a.k.a., DeepMDP~\citep{gelada2019deepmdp}}
This is the same as the forward raw model, only that $f$ now describes a distribution over next state representations. Thus, the log-probability with respect to $f$ becomes $-\log P(\phi(s_{t+1}) | f(\phi(s_t),a_t))$.

\paragraph{Forward energy model} This is the same as the forward raw model, only that $f$ is no longer a distribution over raw states. Rather, $f$ maps $\phi(s_t),a_t$ to the same embedding space as $\phi$ and the probability $P(s_{t+1} | f(\phi(s_t),a_t))$ is defined in an energy-based way:
\begin{equation}
    \frac{\rho(s_{t+1})\exp\{\phi(s_{t+1})^\top W f(\phi(s_t),a_t)\}}{\E_\rho [\exp\{\phi(\tilde{s})^\top W f(\phi(s_t),a_t)\}]},
\end{equation}
where $W$ is a trainable matrix and $\rho$ is a non-trainable prior distribution (we set $\rho$ to be the distribution of states in the offline dataset).

\paragraph{(Momentum) temporal contrastive learning (TCL)}
Given a sub-trajectory $\tau_{t:t+1}$, we apply a contrastive loss between $\phi(s_t),\phi(s_{t+1})$. The objective is
\begin{equation}
    -\phi(s_{t+1})^\top W \phi(s_t) + \log \E_\rho [\exp\{\phi(\tilde{s})^\top W \phi(s_t)\}],
\end{equation}
where $W$ and $\rho$ are as in the forward energy model above. This objective may be generalized to sequences longer than $k+1=2$ by having multiple terms in the loss for $i=1,\dots,k$:
\begin{equation}
-\phi(s_{t+i})^\top W_i \phi(s_t) + \log \E_\rho [\exp\{\phi(\tilde{s})^\top W_i \phi(s_t)\}].
\end{equation}
If momentum is used, we apply the contrastive loss between $f(\phi(s_t))$ and $\phi_{target}(s_{t+i})$, where $f$ is a learned function and $\phi_{target}$ denotes a non-trainable version of $\phi$, with weights corresponding to a slowly moving average of the weights of $\phi$, as in~\citet{stooke2020decoupling,he2020momentum}.

\paragraph{Attentive Contrastive Learning (ACL)} 
Following the theme of contrastive losses and inspired by a number of works in the RL~\citep{oord2019representation} and NLP~\citep{mikolov2013efficient} literature which apply such losses between tokens and contexts using an attention mechanism, we devise a similar objective for our settings.
Implementation-wise, we borrow ideas from BERT~\citep{devlin2018bert}, namely we (1) take a sub-trajectory $s_{t:t+k},a_{t:t+k},r_{t:t+k}$, (2) randomly mask a subset of these, (3) pass the masked sequence into a transformer, and then (4) for each masked input state, apply a contrastive loss between its representation $\phi(s)$ and the transformer output at its sequential position.
We use $k+1=8$ in our implementation.
\figref{fig:framework} provides a diagram of ACL.

\paragraph{Value prediction network (VPN)} Taken from~\citet{oh2017value}, this objective uses an RNN starting at $\phi(s_t)$ and inputting $a_{t:t+k}$ for $k$ steps to predict the $k$-step future rewards and value functions. While the original VPN paper defines the $(k+1)$-th value function in terms of a max over actions, we avoid this potential extrapolation issue and simply use the $(k+1)$-th action provided in the offline data. As we will elaborate on later, VPN bears similarities to ACL in that it uses certain components of the input sequence (states and actions) to predict other components (values).

\paragraph{Deep bisimulation for control} This objective is taken from~\citet{zhang2020learning}, where the representation function $\phi$ is learned to respect an L1 distance based on a bisimulation similarity deduced from Bellman backups.

\subsection{Results}
\begin{figure*}[t]
    \centering
    \includegraphics[width=0.9\linewidth]{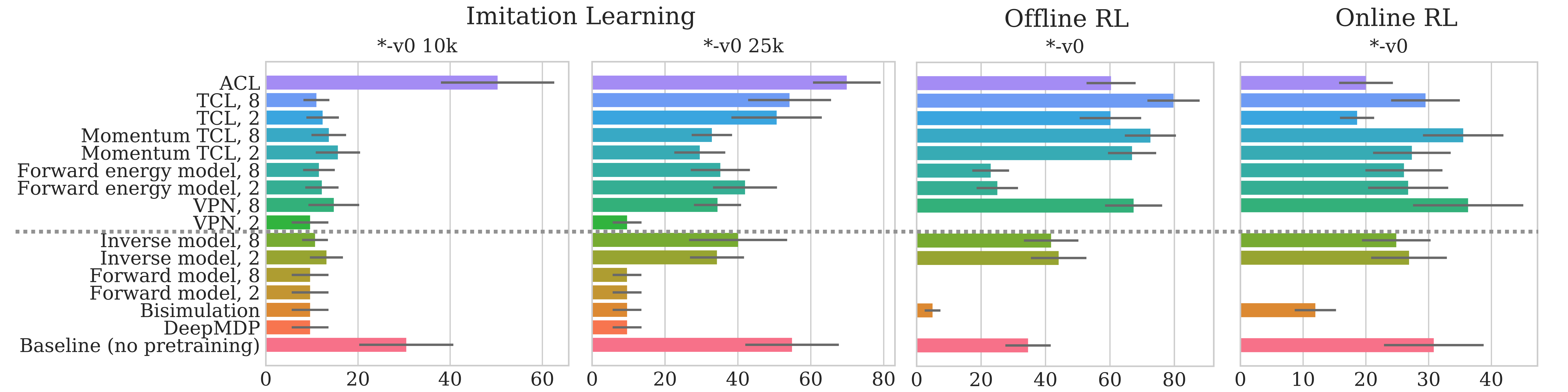}
    \caption{Performance of downstream imitation learning, offline RL, and online RL tasks under a variety of representation learning objectives.  $x$-axis shows aggregated average rewards (over five seeds) across the domains and datasets described in~\secref{sec:setup}. Methods that failed to converge are eliminated from the results (see~\appref{app:exp}). ACL is set to the default configuration that favors imitation learning (see~\secref{exp:depth}). When applicable, we also label variants with $k+1\in\{2,8\}$. Methods above the dotted line are variants of contrastive self-prediction. ACL performs well on imitation learning. VPN and (momentum) TCL perform well on offline and online RL.}
    \label{fig:baselines}
\end{figure*}
The results of these representation learning objectives are presented in~\figref{fig:baselines}. Representation learning, even before the extensive ablations we will embark on in~\secref{exp:depth}, on average improves downstream imitation learning, offline RL, and online RL tasks by $1.5$x, $2.5$x, and $15\%$ respectively. 
The objectives that appear to work best -- ACL, (Momentum) TCL, VPN -- fall under a class of objectives we term \emph{contrastive self-prediction}, where \emph{self-prediction} refers to the idea that certain components of a sub-trajectory are predicted based on other components of the same sub-trajectory, while \emph{contrastive} refers to the fact that this prediction should be performed via a contrastive energy-based loss when the predicted component is a state observation. 

We also find that a longer sub-trajectory $k+1=8$ is generally better than a short one $k+1=2$. The advantage here is presumably due to the non-Markovian nature of the dataset. Even if the environment is Markovian, the use of potentially distinct policies for data collection can lead to non-Markovian data.

Despite these promising successes, there are a number of objectives which perform poorly. Raw predictions of states (forward model) yields disappointing results in these settings. 
Forward models of future representations -- DeepMDP, Bisimulation -- also exhibit poor performance. This latter finding was initially surprising to us, as many theoretical notions of state abstractions are based on the principle of predictability of future state representations. Nevertheless, even after extensive tuning of these objectives and attempts at similar objectives (e.g., we briefly investigated incorporating ideas from~\citet{hafner2020mastering}), we were not able to achieve any better results. Even if it is possible to find better architectures or hyperparameters, we believe the difficulty in tuning these baselines makes them unattractive in comparison to the simpler and better performing alternatives.

\section{Experiments: Depth Study}\label{exp:depth}
\begin{table*}[t]
\footnotesize
\centering
\caption{Factors of contrastive self-prediction considered in our ablation study and summaries of their effects. Input action and input reward default to true. The remaining factors default to false. For each effect entry, $\downarrow$ means decreased performance, $\uparrow$ means improved performance, and $=$ means no significant effect.}
\label{tab:conclusion}
\setlength{\tabcolsep}{3pt}
\renewcommand{\arraystretch}{1.1}
\begin{tabular}{l|p{9cm}|c|c|c}
\toprule
\textbf{Factor} & \textbf{Description} & \textbf{Imitation} & \textbf{Offline} & \textbf{Online} \\\midrule
reconstruct action & Add action prediction loss based on $\phi(s)$. & $\downarrow$ & $\uparrow$ & $\uparrow$  \\\hline
reconstruct reward & Add a reward prediction loss based on $\phi(s)$. & $\downarrow$ & $\uparrow$ & $\uparrow$ \\\hline
predict action & Add an action prediction loss based on transformer outputs. Whenever this is true, we also set `input embed' to true. & $\downarrow$ & $\uparrow$ & $\uparrow$ \\\hline
predict reward & Add a reward prediction loss based on transformer outputs. Whenever this is true, we also set `input embed' to true. & $\downarrow$ & $\uparrow$ & $\uparrow$ \\\hline
input action & Include actions in the input sequence to transformer. & $\downarrow$ & $\uparrow$ & $\uparrow$  \\\hline
input reward & Include rewards in the input sequence to transformer. & $\downarrow$ & $\uparrow$ & $\uparrow$ \\\hline
input embed & Use representations $\phi(s)$ as input to transformer, as opposed to raw observations. & $\downarrow$ & $=$ & $\uparrow$ \\\hline
bidirectional & To generate sequence output at position $i$, use full input sequence as opposed to only inputs at position $\le i$. & $\downarrow$ & $=$ & $\uparrow$ \\\hline
finetune & Pass gradients into $\phi$ during learning on downstream tasks. & $\downarrow$ & $\downarrow$ & $\uparrow$ \\\hline
auxiliary loss & Use representation learning objective as an auxiliary loss during downstream learning, as opposed to pretraining. & $\downarrow$ & $\downarrow$ & $\uparrow$ \\\hline
momentum & Adopt an additional momentum representation network. Whenever this is true, we also set `input embed' to true. & $\downarrow$ & $\downarrow$ & $\uparrow$ \\\hline
discrete embedding & Learn discrete representations. Following~\citet{hafner2020mastering}, we treat the 256-dim output of $\phi$ as logits to sample 16 categorical distributions of dimension 16 each and use  straight-through gradients. & $\downarrow$ & $\downarrow$ & $\downarrow$ \\\hline
context embedding & Following~\citet{devlin2018bert}, use transformer output as representations for downstream tasks. Whenever this is true, we also set `input embed' to true. & $\downarrow$ & $\downarrow$ & $\downarrow$ \\
\bottomrule
\end{tabular}
\end{table*}

\begin{figure*}[t!]
    \centering
    \includegraphics[width=0.95\linewidth]{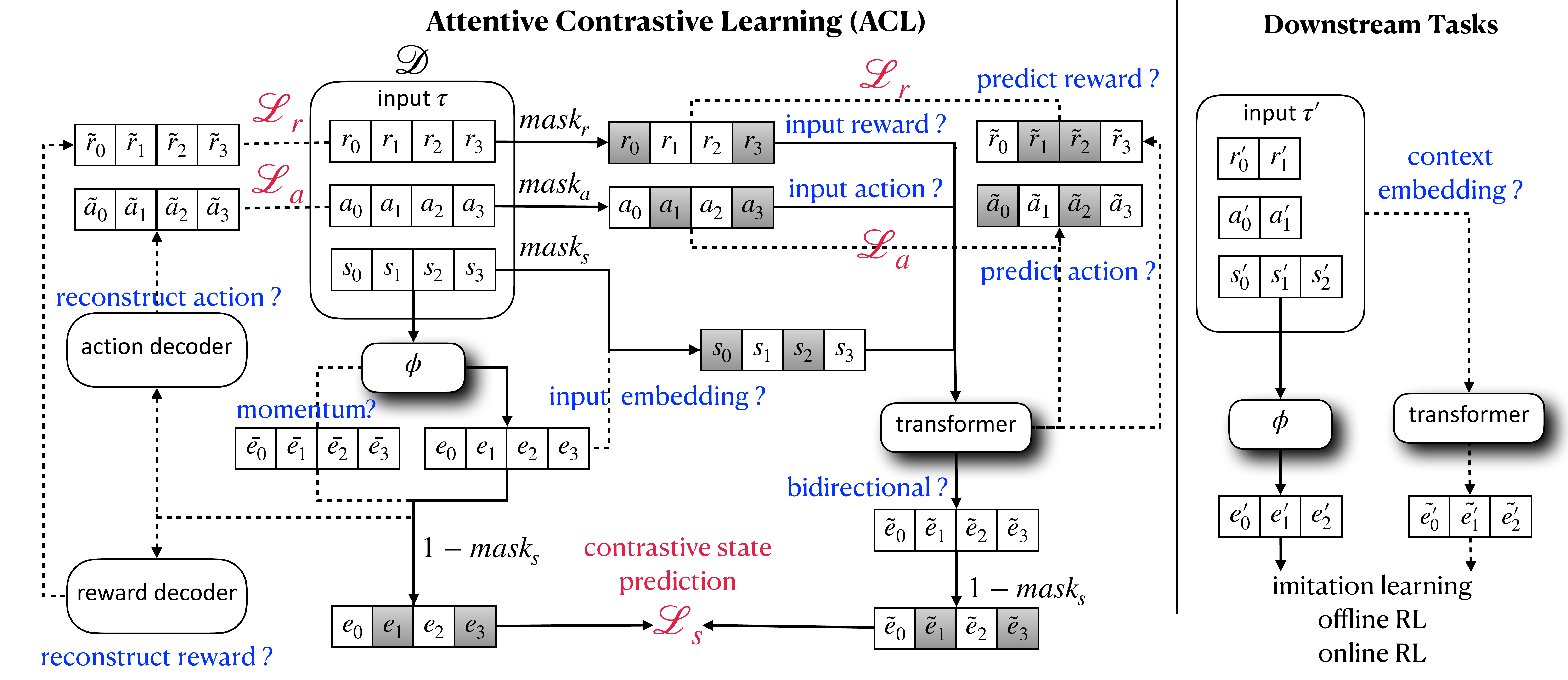}
    \caption{A pictoral representation of our depth study based on contrastive self-prediction. We use the transformer-based architecture of attentive contrastive learning (ACL) as a skeleton for ablations with respect to various representation learning details. Solid arrows correspond to the configuration of ACL. Dotted arrows and blue text are factors considered in the ablation study. Gray blocks are masked state/action/reward entries. After the pretraining phase, the representation network $\phi$ is reused for downstream tasks, unless `context embedding' is true, in which case the transformer is used.
    }
    \label{fig:framework}
\end{figure*}

\begin{figure*}[t]
    \centering
    \includegraphics[width=0.9\linewidth]{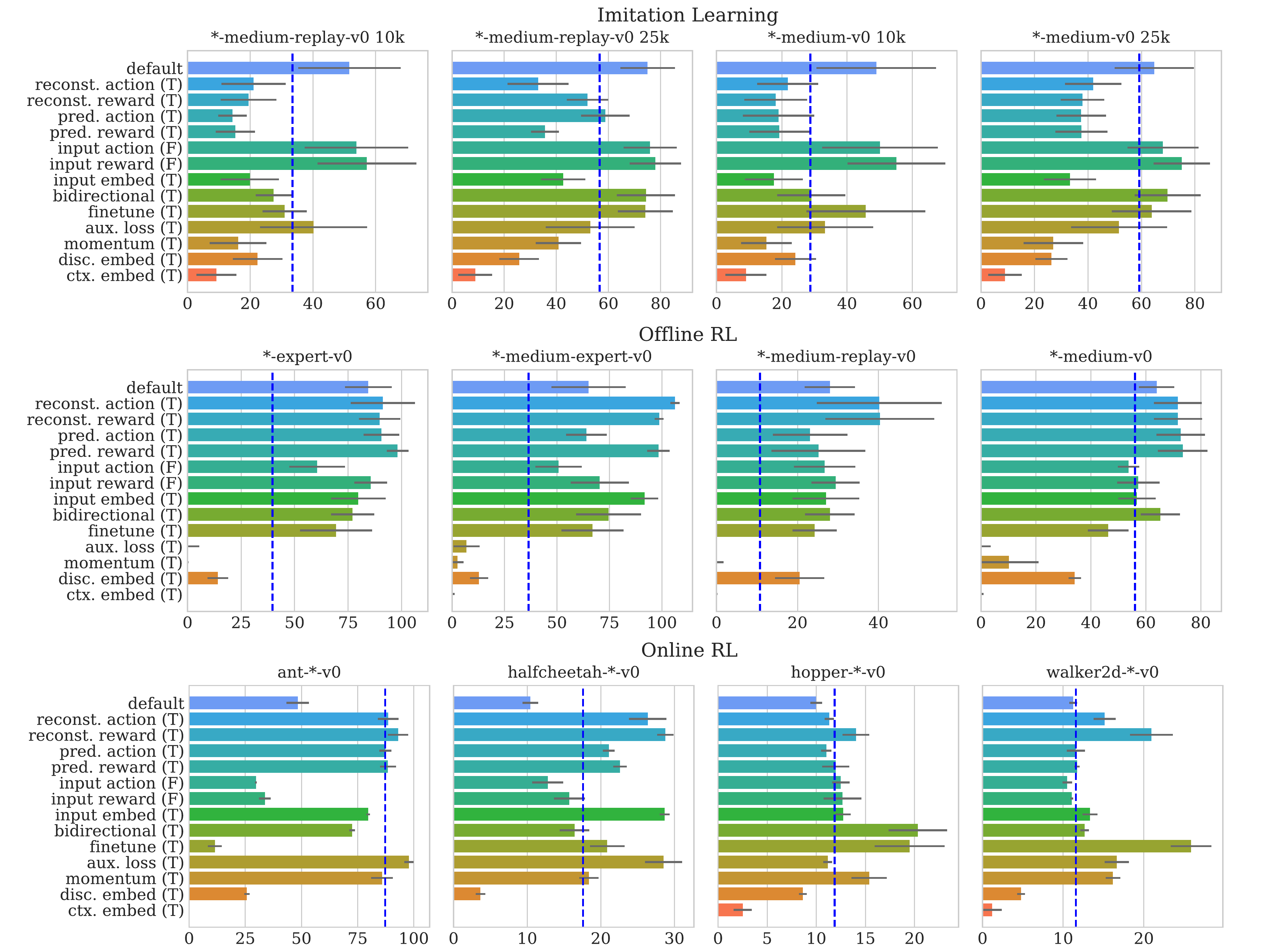}
    \caption{Ablation results on imitation learning, offline RL, and online RL. $x$-axis shows average rewards and standard error aggregated over either different Gym-MuJoCo datasets (imitation and offline RL) or domains (online RL). Blue dotted lines show average rewards without pretraining. (T) and (F) mean setting each factor to true or false (opposite from the default configuration). Reconstructing, predicting, or inputting action or reward (row 2-7) impairs imitation performance but are important for offline and online RL. Bidirectional transformer hurts imitation learning when downstream sample size is small. Finetuning and auxiliary loss can help online RL. Additional results are presented in~\appref{app:exp_results}.}
    \label{fig:ablation}
\end{figure*}

The favorable results of objectives based on the idea of contrastive self-prediction is compelling, but the small number of objectives evaluated leaves many questions unanswered.
For example, when generating the context embedding for a specific prediction, should one use past states (as in TCL and Momentum TCL) or also include actions and/or rewards (as in ACL and VPN)? Should this context use the same representation network $\phi$ (as in TCL and VPN), a momentum version of it (as in Momentum TCL), or a completely separate network (as in ACL)?

We use this section to study these and other important questions by conducting a series of ablations on the factors which compose a specific contrastive self-prediction objective and how it is applied to downstream learning. We describe all these factors in~\tabref{tab:conclusion}, as well as a high-level summary of their effects. 
Further anecdotal observations found during our research are summarized in~\appref{app:anecdotal}.

We choose the transformer-based implementation of ACL to serve as the skeleton for all these ablations (see~\figref{fig:framework}), due to its general favorable empirical performance in the previous section, as well as its ease of modification.
For each downstream task below, we present the ablations with respect to the \emph{default} configuration of the factors in~\tabref{tab:conclusion} that corresponds to the original ACL introduced in~\secref{sec:breadth}, and change one factor at a time to observe its effect on downstream task performance.

\subsection{Results}
The results of our ablation studies are presented in Figure~\ref{fig:ablation}, and we highlight some of the main findings below. We also take the best performing ablation from each row (imitation, offline RL, and online RL) and plot the performance during training in~\figref{fig:line_plot}.

Let us first consider the effects of inclusion or prediction of actions and rewards. We notice some interesting behavior across the different downstream modes. Namely, it appears that imitation learning is best served by focusing only on state contrastive learning and not including or predicting actions and rewards, whereas the offline and online RL settings appear to benefit from these.
Due to the mixed results we initially observed from including or predicting actions and rewards, we also introduce the idea of \emph{reconstructing} actions and rewards based on $\phi(s)$, and we found this to have much more consistent benefit in the RL settings, although it still degrades imitation learning performance. 
This disconnect between objectives which are good for imitation learning vs. RL, first seen in~\secref{sec:breadth}, thus continues to be present in these ablations as well, and we find that no single objective dominates in all settings. 

We also evaluate a number of representation learning paradigms popular in the NLP literature~\citep{devlin2018bert}, namely using bidirectional transformers, finetuning, and context embedding.
Although these techniques are ubiquitous in the NLP literature, we find mixed results in RL settings. Context embedding consistently hurts performance. Bidirectional transformer hurts imitation learning but helps online RL. Finetuning leads to a modest degredation in performace in imitation and offline RL but can improve online RL depending on the domain being evaluated.

We additionally considered using the representation learning objective as an auxiliary training loss, which is popular in the online RL literature~\citep{ShelhamerMAD16,stooke2020decoupling}.
And indeed, we find that it can dramatically improve representation learning in online RL, but at the same time, dramatically degrade performance in the offline settings (imitation learning or offline RL).

%% file: conc.tex
% !TEX root = main.tex
\section{Conclusion}
Overall, our results show that relatively simple representation learning objectives can dramatically improve downstream imitation learning, offline RL, and online RL (\figref{fig:line_plot}).
Interestingly, our results suggest that the ideal representation learning objective may depend on the nature of the downstream task, and no single objective appears to dominate generally.
Our extensive ablations also provide a number of intriguing insights, showing that representational paradigms which are popular in NLP or online RL may not translate to good performance in offline settings.

Even with this multitude of fresh insight into the question of representation learning in RL, our study is limited in a number of aspects, and these aspects can serve as a starting point for future work. For example, one may consider additional downstream tasks such as multi-task, transfer, or exploration. Alternatively, one can extend our ablations to real-world domains like robot learning. Or, one may consider ablating over different network architectures.

Despite these limitations, we hope our current work proves useful to RL researchers, and serves as a guide for developing even better and more general representation learning objectives.

%% file: appendix.tex
% !TEX root = main.tex
\clearpage
\newpage

\appendix
\onecolumn

\begin{appendix}

\thispagestyle{plain}
\begin{center}
{\huge Appendix}
\end{center}

\section{Experimental Details}\label{app:exp}
\subsection{Representation Network}
\begin{figure}[h]
    \centering
    \includegraphics[width=0.9\linewidth]{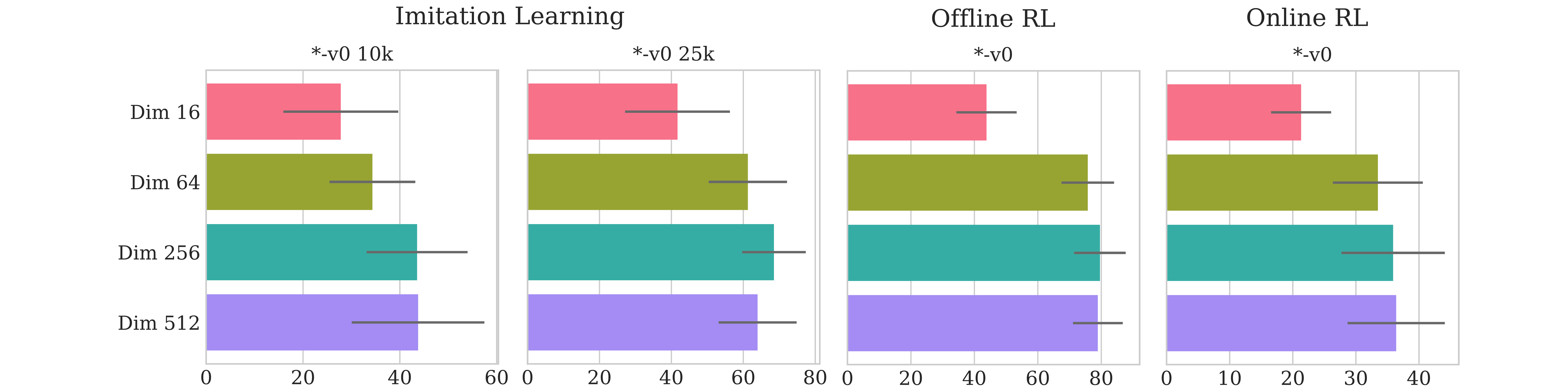}
    \caption{Average reward across domains and datasets with different representation dimensions. $256$ and $512$ work the best (this ablation is conducted with ``reconstruct action'' and ``reconstruct reward'' set to true).}
    \label{fig:app_dim}
\end{figure}

We parametrize the representation function $\phi$ as a two-hidden layer fully connected neural network with $256$ units per layer and output dimension $256$. A Swish~\citep{ramachandran2017searching} activation function is applied to the output of each hidden layer.  We experimented with representation dimension sizes $16, 64, 256,$ and $512$, and found $256$ and $512$ to generally work the best (see \figref{fig:app_dim} in \appref{app:repr}).

\subsection{Transformer Network} The BERT-style transformer used in attentive contrastive learning (ACL) consists of one preprocessing layer of $256$ units and ReLU activaiton, followed by a multi-headed attention layer ($4$ heads with $128$ units each), followed by a fully connected feed forward layer with hidden dimension $256$ and ReLU activation, finally followed by an output layer of $256$ units (the same as $\phi$'s output). We experimented with different number of attention blocks and number of heads in each block, but did not observe significant difference in performance.

When masking input (sequences of state, actions, or rewards), we randomly choose to `drop' each item with probability $0.3$, 'switch' with probability $0.15$, and `keep' with probability $0.15$. 
`Drop' refers to replacing the item with a trainable variable of the same dimension.
`Switch' refers to replacing the item with a randomly sampled item from the same batch.
`Keep' refers to leaving the item unchanged.
These probability rates where chosen arbitrarily and not tuned.

\subsection{Action Prediction and Reconstruction}
Whenever a loss includes action prediction or reconstruction, we follow~\citet{haarnoja2019soft}, and (1) utilize an output distribution given by a tanh-squashed Gaussian and (2) apply an additive adaptive entropy regularizer to the action prediction loss.

\subsection{Other Networks}
With few exceptions, all other functions $f,g$ mentioned in~\secref{sec:breadth} are two-hidden layer fully connected neural networks with $256$ units per layer and using a Swish~\citep{ramachandran2017searching} activation.

The only exception is Momentum TCL, where $f$ is the same structure but using a residual connection on the output.

\subsection{Training}
During pretraining, we use the Adam optimizer with learning rate $0.0001$, except for the TCL variants, for which we found $0.0003$ to work better.
For Momentum TCL, we use a moving average with rate $0.05$.

\subsection{Convergence Failures}
Representations learned under objectives including forward-raw model, VPN (with $k+1=2$), and DeepMDP consistently diverge and output NaNs on offline and online RL, and are therefore removed from the results in~\figref{fig:baselines}. The bisimulation objective on offline and online RL fails to converge in some runs but occasionally succeeds, therefore the means of succeeded runs are computed and shown in~\figref{fig:baselines}.

\clearpage
\section{Additional Experimental Results}\label{app:exp_results}

\subsection{Ablation Over Representation Size}
\label{app:repr}

\subsection{Additional Contrastive Learning Results}
\begin{figure}[h]
    \centering
    \includegraphics[width=.8\linewidth]{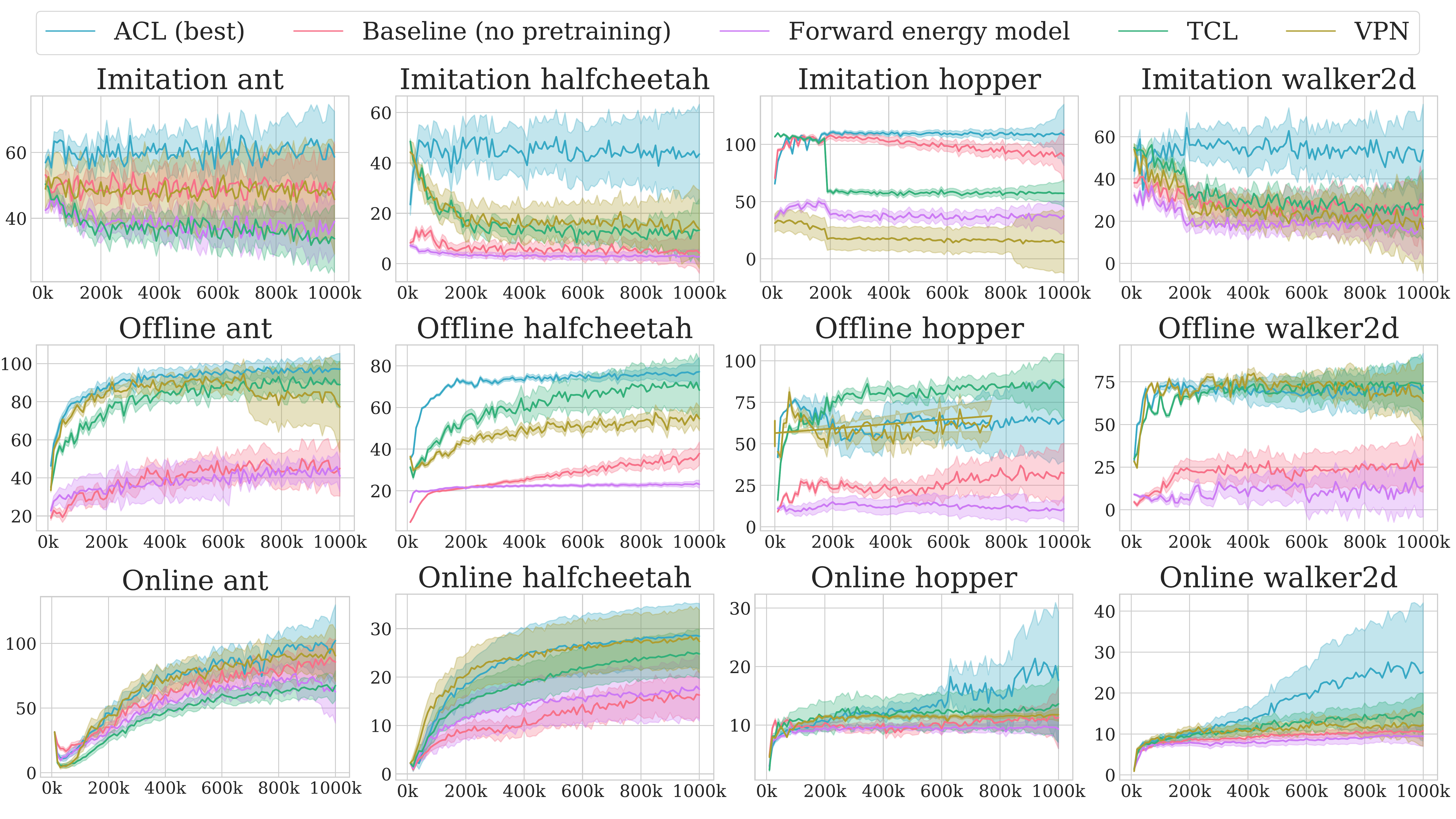}
    \caption{Additional training curves of contrastive learning objectives aggregated over different offline datasets in the same domain. Both in this figure and in~\figref{fig:line_plot}, we plot the best variant of ACL according to the ablation study, namely we set ``input reward'' to false in imitation learning, ``reconstruct action'' to true in offline RL, and ``auxiliary loss'' (in ant and halfcheetah) or ``finetuning'' (in hopper and walker2d) to true in online RL. The best variant of ACL generally performs the best compared to other contrastive learning objectives, although TCL's performance is competitive in offline RL.}
    \label{fig:baseline_lineplot}
\end{figure}

\clearpage
\subsection{Ablation Results for Individual Domains and Datasets}
\begin{figure}[h]
    \centering
    \includegraphics[width=.9\linewidth]{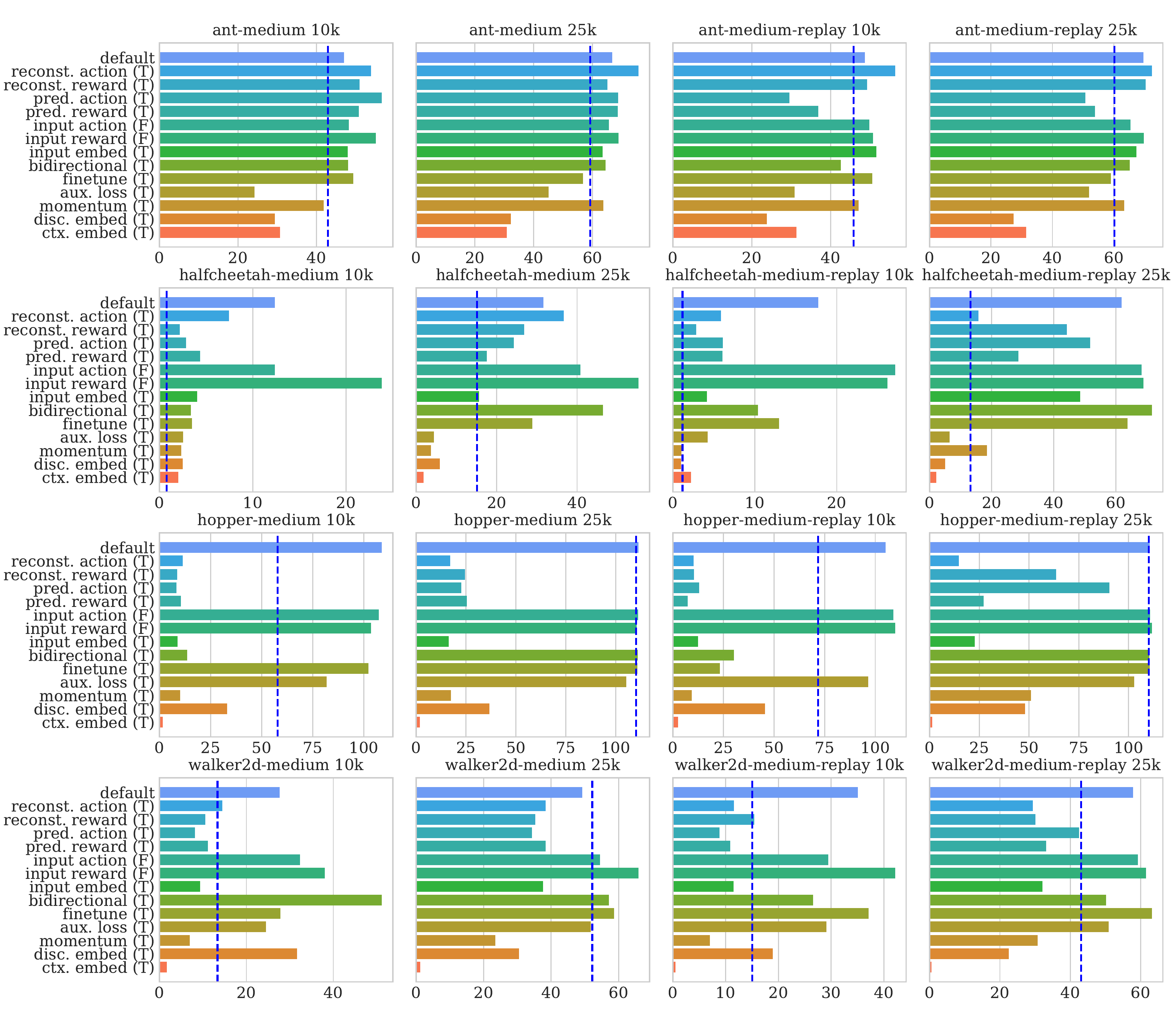}
    \caption{Imitation learning ablation on individual domains and datasets. The negative impact of inputting action and reward to pretraining is more evident in halfcheetah and walker2d. Reconstructing/predicting action/reward is especially harmful in halfcheetah, hopper, and walker2d. There always exists some variant of ACL that is better than without representation learning (blue lines) in all domain-dataset combinations.}
    \label{fig:bc_none}
\end{figure}
\begin{figure}[h]
    \centering
    \includegraphics[width=.9\linewidth]{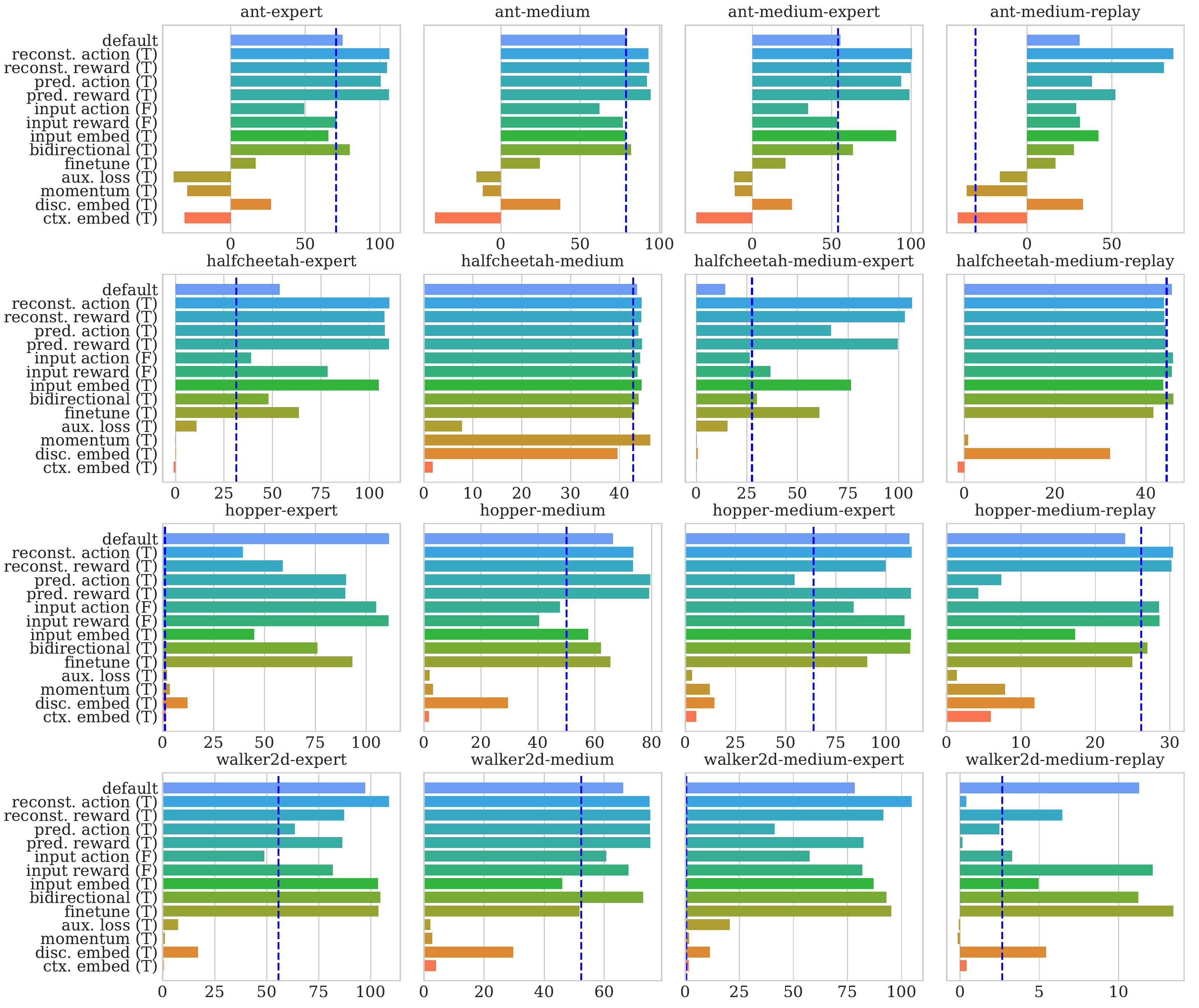}
    \caption{Offline RL ablation on individual domains and datasets. The benefit of representation learning is more evident when expert trajectories are present (e.g., expert and medium-expert) than when they are absent (medium and medum-replay). Reconstructing action and reward is more important in ant and halfcheetah than in hopper and walker2d.} 
    \label{fig:brac_none}
\end{figure}
\begin{figure}[h]
    \centering
    \includegraphics[width=.9\linewidth]{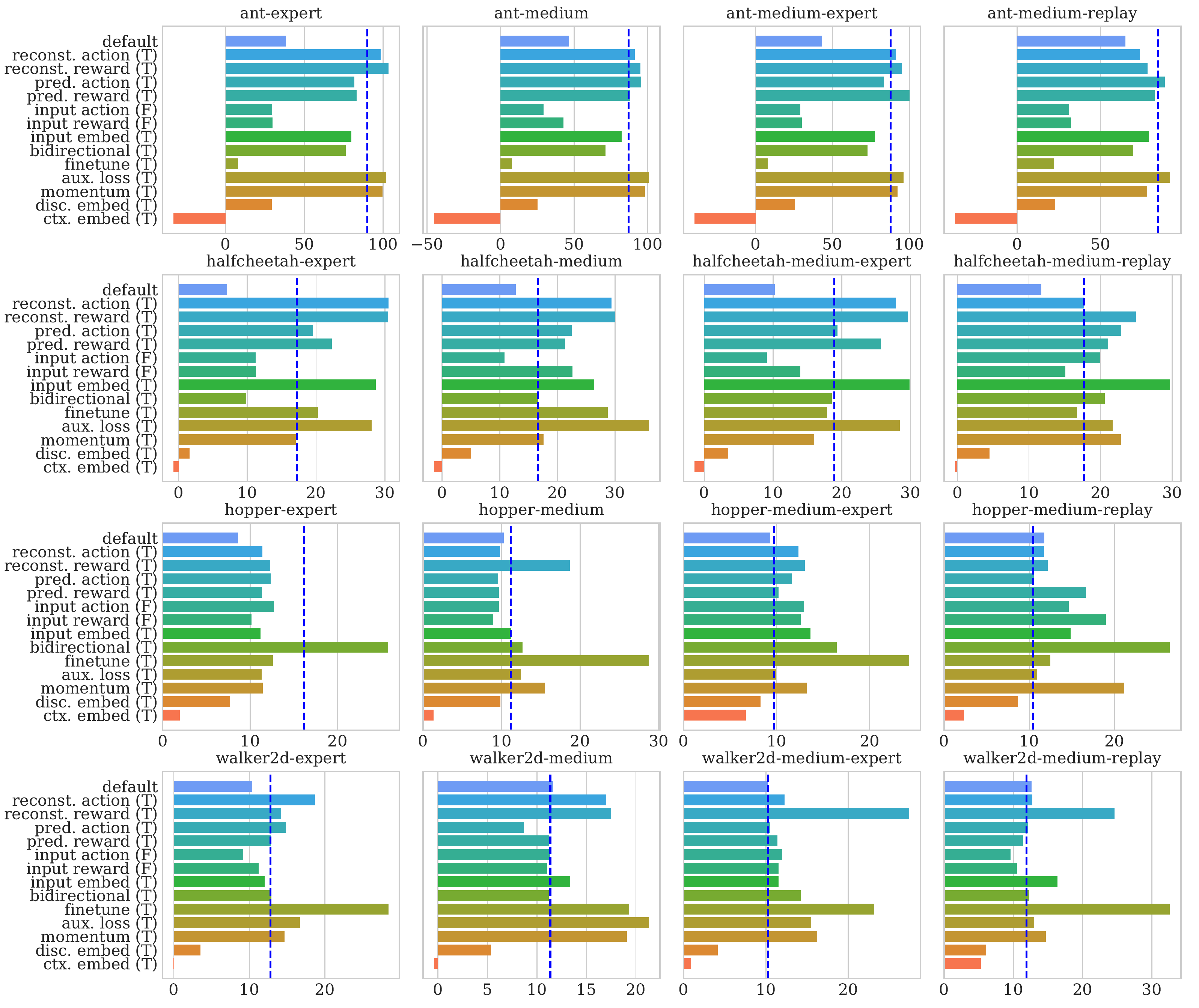}
    \caption{Online RL ablation on individual domains and datasets. Auxiliary loss generally improves performance in all domains and datasets. Finetuning improves halfcheetah, hopper, and walker2d but significantly impairs ant.}
    \label{fig:sac_none}
\end{figure}

\clearpage
\section{Additional Anecdotal Conclusions}\label{app:anecdotal}

\begin{enumerate}
    \item {\bf More ablations.} Although we present our ablations as only changing one factor at a time, we also experimented with changing multiple factors at a time. We did not find any of these additional ablations to change the overall conclusions.
    \item {\bf Reconstruct action.} One ablation that did work surprisingly well was to only reconstruct the action (with no other loss). This appeared to perform well on imitation learning, but poorly on other settings. 
    \item {\bf More transformers.} We experimented with a different application of transformers than ACL. Namely, we attempted to treat each dimension of the state as a token in a sequence (as opposed to using the whole state observation as the token). We found this to provide promising results, although it did not convincingly improve upon the configuration of ACL. Still, it may merit further investigation by future work.
    \item {\bf Transformer architecture.}  We experimented with a different number of attention blocks or number of heads in each block, but did not observe significant differences in performance.
    \item {\bf Normalized or regularized representations.} We experimented with applying an explicit normalization layer on the output of $\phi$ and found no benefits. We also experimented with a stochastic representation along with a KL-divergence regularizer to the standard normal distribution, and again found no benefits.
    
\end{enumerate}
\end{appendix}